\documentclass[a4paper, 10pt, conference]{ieeeconf}      

\IEEEoverridecommandlockouts                              

\usepackage{graphics} 
\usepackage{epsfig} 
\usepackage{amsmath}
\usepackage{pgfplots}
\usepackage{color}
\usepackage{subfigure}
\usepackage[ruled,vlined]{algorithm2e}
\usepackage{amsmath}
\usepackage[flushleft]{threeparttable}
\usepackage{hyperref}

\DeclareMathOperator*{\argmin}{\arg\!\min}

\title{\LARGE \bf
\textit{LATTE}: Accelerating LiDAR Point Cloud Annotation via Sensor Fusion, One-Click Annotation, and Tracking
}

\author{Bernie Wang$^{1}$, Virginia Wu$^{2}$, Bichen Wu$^{1}$, and Kurt Keutzer$^{1}$
\thanks{$^{1}$Bernie Wang, Bichen Wu, and Kurt Keutzer are with Berkeley AI Research, University of California, Berkeley, CA, U.S.A. Email: {\tt\small \{berniewang, bichen, keutzer\}@berkeley.edu}
        }%
\thanks{$^{2}$Virginia Wu is with the Department of Software Engineering, Huaqiao University, Fujian Province, China. She worked on this project while interning at UC Berkeley. Email: {\tt\small wbc0310@gmail.com}}%
}

\begin{document}

\maketitle
\thispagestyle{empty}
\pagestyle{empty}

\begin{abstract}
LiDAR (Light Detection And Ranging) is an essential and widely adopted sensor for autonomous vehicles, particularly for those vehicles operating at higher levels (L4-L5) of autonomy. Recent work has demonstrated the promise of deep-learning approaches for LiDAR-based detection. However, deep-learning algorithms are extremely data hungry, requiring large amounts of labeled point-cloud data for training and evaluation. Annotating LiDAR point cloud data is challenging due to the following issues: 1) A LiDAR point cloud is usually sparse and has low resolution, making it difficult for human annotators to recognize objects. 2) Compared to annotation on 2D images, the operation of drawing 3D bounding boxes or even point-wise labels on LiDAR point clouds is more complex and time-consuming. 3) LiDAR data are usually collected in sequences, so consecutive frames are highly correlated, leading to repeated annotations. To tackle these challenges, we propose \textit{LATTE}, an open-sourced annotation tool for LiDAR point clouds. LATTE features the following innovations: 1) Sensor fusion: We utilize image-based detection algorithms to automatically pre-label a calibrated image, and transfer the labels to the point cloud. 2) One-click annotation: Instead of drawing 3D bounding boxes or point-wise labels, we simplify the annotation to just one click on the target object, and automatically generate the bounding box for the target. 3) Tracking: we integrate tracking into sequence annotation such that we can transfer labels from one frame to subsequent ones and therefore significantly reduce repeated labeling. Experiments show the proposed features accelerate the annotation speed by 6.2x and significantly improve label quality with 23.6\% and 2.2\% higher instance-level precision and recall, and 2.0\% higher bounding box IoU. LATTE is open-sourced at \url{https://github.com/bernwang/latte}. 
\end{abstract}

\section{Introduction}
LiDAR (Light detection and ranging) is an essential and widely adopted sensor for autonomous vehicles. This is particularly true for applications such as Robo-Taxis which require higher levels (L4-L5) of autonomy. Compared with cameras, LiDAR is more robust to ambient light condition changes. It can also provide very accurate  distance measurements (error $<2$cm $^1$) to nearby obstacles, which is essential for the planning and control of autonomous vehicles. 
\begin{figure}[h]
    \centering
    \includegraphics[width=0.9\linewidth]{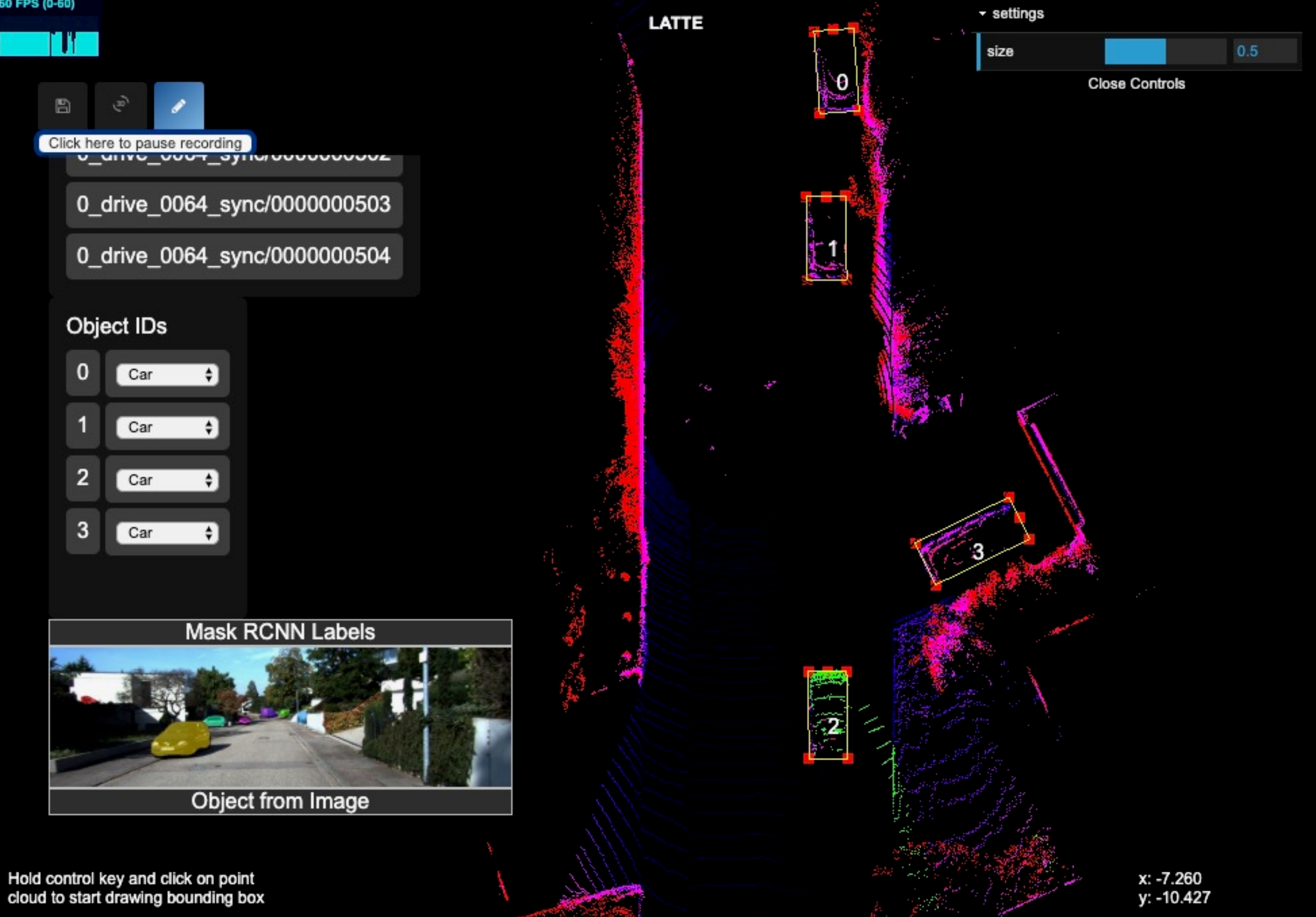}
    \caption{A screenshot of LATTE. Best viewed in color.}
    \label{fig:abstract}
\end{figure}

To understand the environment through LiDAR, autonomous vehicles need to extract semantic meaning from the point cloud and accurately identify and locate objects such as cars, pedestrians, cyclists, and so on. Such problems are called LiDAR-based detection, and they have long been studied by the research community. An increasing number of works \cite{wu2018squeezeseg, wu2018squeezesegv2,qi2017frustum,LiDARDet} have demonstrated the promise of using Deep Learning to solve this problem. Compared with previous approaches, deep learning solutions obtain superior accuracy and faster speed, but they are also extremely data hungry, requiring large amounts of data for training.  

\begin{figure*}[h]
    \centering
    \includegraphics[width=0.8\textwidth]{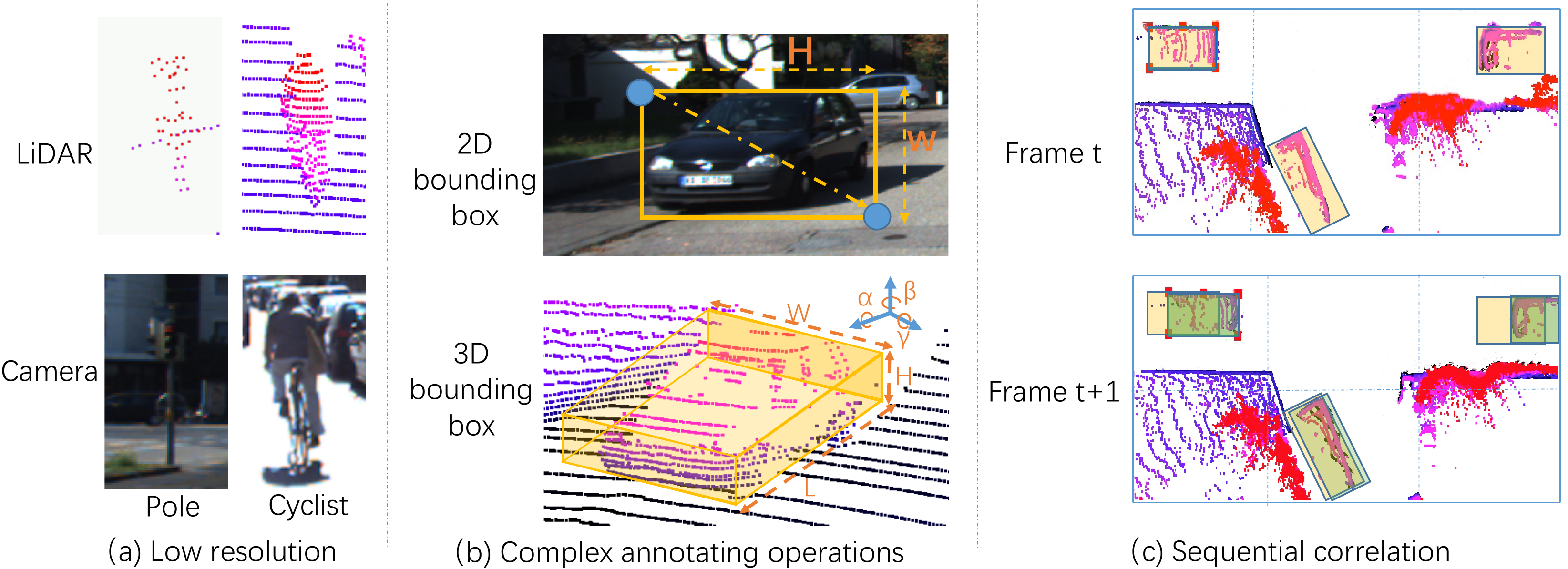}
    \caption{Challenges of annotating LiDAR point clouds. (a) LiDAR point clouds have low resolution and therefore objects are difficult for humans to recognize. The upper two figures are point clouds of a traffic pole and a cyclist, but both are difficult to recognize. The lower two are the corresponding images. (b) Annotating 2D bounding boxes on an image vs. 3D bounding boxes on a point cloud. Annotating 3D bounding boxes is more complicated due to more degrees of freedom of 3D scaling and rotation. (c) Point clouds of two consecutive frames are shown here. Even though the two frames are highly similar, target objects are moving and have different speeds. As a result, new bounding boxes are needed on new frames.}
    \label{fig:challenges}
\end{figure*}

Compared to annotating camera images, annotating LiDAR point clouds is much more difficult. The challenges can be summarized in the following three aspects and are illustrated in Fig. \ref{fig:challenges}: 1) \textbf{Low resolution}: Cameras can easily capture high-resolution images, with widespread support for 4K resolution. In comparison, LiDAR sensors have much more limited resolution. For example, typical vertical resolutions for Velodyne LiDARs are 32 or 64 lines. For the 64-line LiDAR\footnote{\url{https://velodynelidar.com/hdl-64e.html}} with a vertical angular resolution of 0.41$^\circ$, the spatial resolution at 50 meters is merely 0.36 meters. As a result, LiDAR point clouds are very sparse, making it difficult for human annotators to identify objects, as shown in Fig. \ref{fig:challenges}(a). 2) \textbf{Complex annotating operations}: LiDAR-based detection problems have different formulations, and the most popular two are 3D bounding box detection and point-wise segmentation. The former requires predicting a 3D bounding box that tightly covers a target object, and the latter requires finding all the points that belong to a target object. In both scenarios, annotating 3D point cloud is significantly more complex than annotating a 2D image. In bounding box detection, for example, a 2D bounding box can be determined by drawing two corners. For 3D bounding boxes, however, annotators have to determine not only the center position, the length, width, and height of the target, but also 3D rotations. As a result, correctly annotating a 3D bounding box is much more complex for human annotators, as shown in Fig. \ref{fig:challenges}(b). 3) \textbf{Sequential correlation}: Many LiDAR point cloud data are collected in sequences, so consecutive frames are different but highly correlated. If we were to annotate LiDAR point cloud frame by frame naively, most of the annotations would be repeated, as shown in Fig. \ref{fig:challenges}(c).

Without addressing these challenges, it is difficult to annotate LiDAR data efficiently over a large dataset. This limits the progress of research for LiDAR-based detection. Furthermore, although several efforts are trying to create more open-sourced datasets (\cite{KITTI,huang2018apolloscape}) for LiDAR-based detection, the annotation tools behind these datasets are not publicly accessible. Moreover, there are obvious advantages to enabling groups to efficiently annotate LiDAR datasets for their own LiDAR sensors, configurations, and so on.

To address these problems, we propose \textit{LATTE}, an open-sourced LiDAR annotation tool, as shown in Fig. \ref{fig:abstract}. We address the challenges above with the following solutions: 1) \textbf{Sensor fusion}: Cameras have much higher resolution than LiDAR sensors, and image-based detection algorithms are much more mature than LiDAR-based.  LiDAR sensors are usually paired with cameras, and the two sensors are calibrated such that each point from the cloud can be projected to a corresponding pixel in the image. Therefore, we can apply camera-based image detection algorithms and transfer labels from an image to a 3D point cloud. The algorithm-generated labels are not perfect and are limited by the algorithm accuracy, projection and synchronization errors, but we can use them as pre-labels to help human annotators recognize objects and ``fine-tune'' the labels. 2) \textbf{One-click annotation}: We simplify the operational complexity for LiDAR annotation from drawing point-wise labels to drawing 3D bounding boxes, then to top-view 2D bounding boxes, and eventually to one-click annotations. For a target object, a human annotator need only click on one point on it, and we utilize clustering algorithms to expand the point-annotation to the entire object, and automatically estimate a top-view 2D bounding box around the object. From the top-view of a 2D bounding box, we can infer point-wise labels for segmentation problems, simply by treating each point inside the bounding box as part of the target object. 3) \textbf{Tracking}: To reduce repeated annotations on consecutive frames in a sequence, we utilize tracking algorithms to transfer annotations from one frame to subsequent ones. By integrating all these solutions, our annotation tool enables a 6.2x reduction in annotation time while delivering better label quality, as measured by 23.6\% and 2.2\% higher instance-level precision and recall, and 2.0\% higher bounding box IoU. Furthermore, we open-source our annotation tool and build it in a modular way such that each component can be replaced and improved easily if more advanced algorithms for each part become available.

\section{Related Work}
\label{sec:related work}
\textbf{LiDAR-based detection and datasets:} LiDAR-based detection aims to identify and locate objects of interest from a LiDAR point cloud. Two main problem formulations for these point clouds are bounding box object detection \cite{LiDARDet} of which the goal is to draw a tight bounding box that covers the target object; and semantic segmentation \cite{wu2018squeezeseg}, of which the goal is to predict labels for each point in the cloud and therefore find the cluster corresponding to target objects. Earlier works mainly rely on handcrafted geometrical features for segmentation and classification \cite{LiDARSegICRA2012,himmelsbach2008lidar,wang2012could}. More recent works adopt deep learning to solve this problem \cite{wu2018squeezeseg,wu2018squeezesegv2,qi2017pointnet,qi2017pointnet++,qi2017frustum, LiDARDet} and achieve significant improvements in accuracy and efficiency.

Deep learning methods require much more data for training, therefore many efforts have focused on creating public datasets for LiDAR-based detection. The KITTI dataset \cite{KITTI} contains about 15,000 frames of 3D bounding box annotations for road-objects. The Apolloscape dataset \cite{huang2018apolloscape} contains 140,000 frames of point-wise background annotation. Public datasets serve as benchmarks to facilitate research, but they are not enough to support product adoption since different configurations of LiDAR sensors, locations, and so on requires creating different datasets. Therefore, it is equally important to provide annotation tools to enable more groups to create their own datasets.

\textbf{Annotation tools:} Many efforts focus on improving annotation tools to generate more data for deep learning training, but most of the annotation tools are focusing on images. VIA \cite{dutta2016via} provides a simple yet powerful webpage-based tool for drawing bounding boxes and polygons on images. Later works such as PolygonRNN \cite{castrejon2017annotating, acuna2018efficient} seek to utilize more advanced algorithms to facilitate and accelerate human annotations. For video annotation, VATIC \cite{vatic} integrates tracking (mainly linear interpolation) to reduce annotating repeated entities in consecutive frames. For autonomous driving applications, Yu et al. propose Scalabel \cite{yu2018bdd100k}, a package of tools that support annotating bounding boxes and semantic masks on images. Few works have focused on building LiDAR annotation tools. The Apolloscape \cite{huang2018apolloscape} dataset's annotation pipeline uses sensor fusion and image-based detection to generate labels through images. But their 3D annotations are for static backgrounds instead of moving objects. \cite{piewak2018boosting} adopts a sensor-fusion strategy to generate LiDAR point cloud labels using image-based detectors, and directly use them to train LiDAR-based detectors. However, the correctness of such labels is limited by the accuracy of the image detector, projection and synchronization error. Moreover, neither \cite{huang2018apolloscape} nor \cite{piewak2018boosting} have open-sourced their annotation tools. 

\textbf{Data collection through simulation:} To sidestep the difficulty of data collection and annotation, many research efforts aim at using simulation to generate LiDAR point cloud data to train neural networks. Yue et al. \cite{yue2018lidar} built a LiDAR simulator on top of the video game GTA-V. The simulated data are then used to train, evaluate, and verify deep learning models \cite{wu2018squeezeseg,wu2018squeezesegv2,yue2018lidar}. Carla \cite{Dosovitskiy17} is an open-source simulator for autonomous driving, and it supports image and LiDAR data generation. However, due to the distribution shift between the simulated data and the real world data, deep learning models trained on simulated data perform poorly on the real world data. Many works aim to close the gap between simulation and the real world by domain adaptation \cite{wu2018squeezesegv2}. Despite some promising progress, domain adaptation remains a challenging problem and the gap has been reduced but not closed. Therefore, collecting and annotating real-world data is still critical.

\section{Method}
\label{sec:method}
In this section, we discuss in detail three features of LATTE that aim to accelerate LiDAR point cloud annotation: sensor fusion, one-click annotation, and tracking.

\subsection{Sensor Fusion}
\begin{figure}[h]
    \centering
    \includegraphics[width=\linewidth]{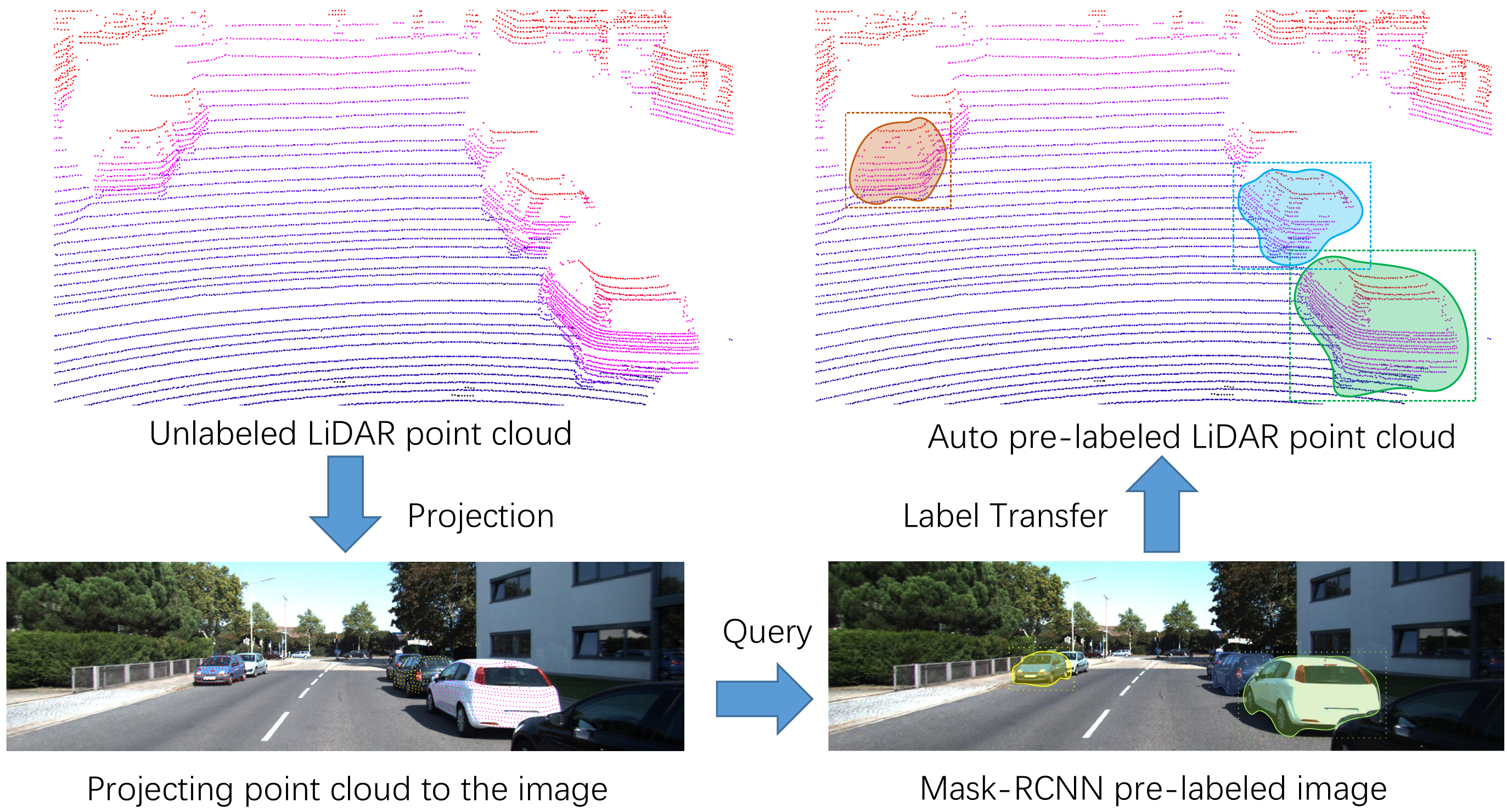}
    \caption{The sensor-fusion pipeline of LATTE. A LiDAR point cloud is projected onto its corresponding image. Next, we use Mask-RCNN to predict semantic labels on the image. The labels are then transferred back to the LiDAR point cloud.}
    \label{fig:sensor-fusion}
\end{figure}

As shown in Fig.\ref{fig:challenges}(a), LiDAR point cloud are low resolution and are difficult for human annotators to recognize. In comparison, cameras have higher resolution, and image-based detection algorithms are more mature than LiDAR-based. Therefore, we use image-based detection to help us annotate LiDAR point cloud. Our pipeline is illustrated in Fig. \ref{fig:sensor-fusion}.

LiDAR sensors are paired with cameras, and the two sensors are calibrated such that for each point $\mathbf{p}_i$ with a 3D coordinate $(x_i, y_i, z_i)$ in the point cloud, we can project it to a pixel $\mathbf{q}_i$ with a 2D coordinate $(u_i, v_i)$ in the corresponding image. The projection can be mathematically described as
\[
\mathbf{q}_i = P\mathbf{p}_i + \mathbf{t},
\]
where $P\in \mathcal{R}^{2\times 3}$ is the projection matrix and $\mathbf{t} \in \mathcal{R}^2$ is the translation vector.  

We then apply semantic segmentation on the image. In our annotator, we use Mask R-CNN \cite{he2017mask} to get semantic labels for each pixel in the image. The semantic labels can be regarded as a mask $\mathcal{M}$ such that for each pixel $\mathbf{q}_i$ with coordinates $(u_i, v_i)$, we can find its label $l_i$ as $l_i = \mathcal{M}(u_i, v_i)$. Finally, this label can be transferred to its corresponding point in the 3D space. This way, we can automatically generate pre-labels for the point cloud. 

\begin{figure}[h]
    \centering
    \includegraphics[width=0.9\linewidth]{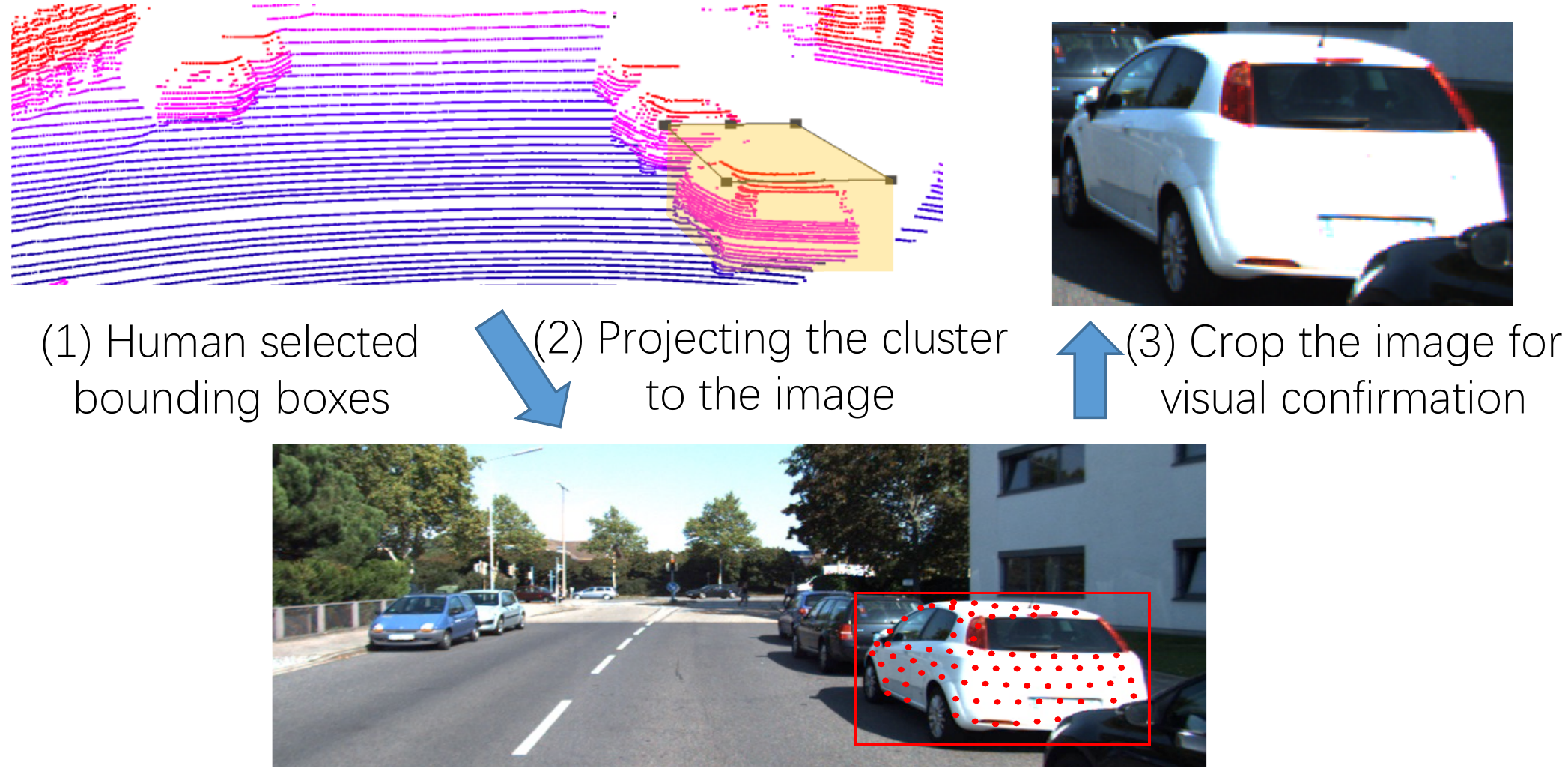}
    \caption{We use sensor fusion to help annotators confirm the category of a selected object. Once a 3D bounding box is chosen, we project all the points within the bounding box to the image and show the corresponding crop of the image to human annotators for visual confirmation.}
    \label{fig:visual-conf}
\end{figure}

We highlight the pre-labeled points in the original point cloud such that human annotators can quickly identify objects of interest. After an annotator draws a bounding box over the target object, we again project all the points in the cluster back to the image, find the patch of the image that contains the target object, crop the patch and show it to human annotators for confirmation, as illustrated in Fig. \ref{fig:visual-conf}. 

\subsection{One-click Annotation}
\label{sec:one-click}
\begin{figure}[h]
    \centering
    \includegraphics[width=\linewidth]{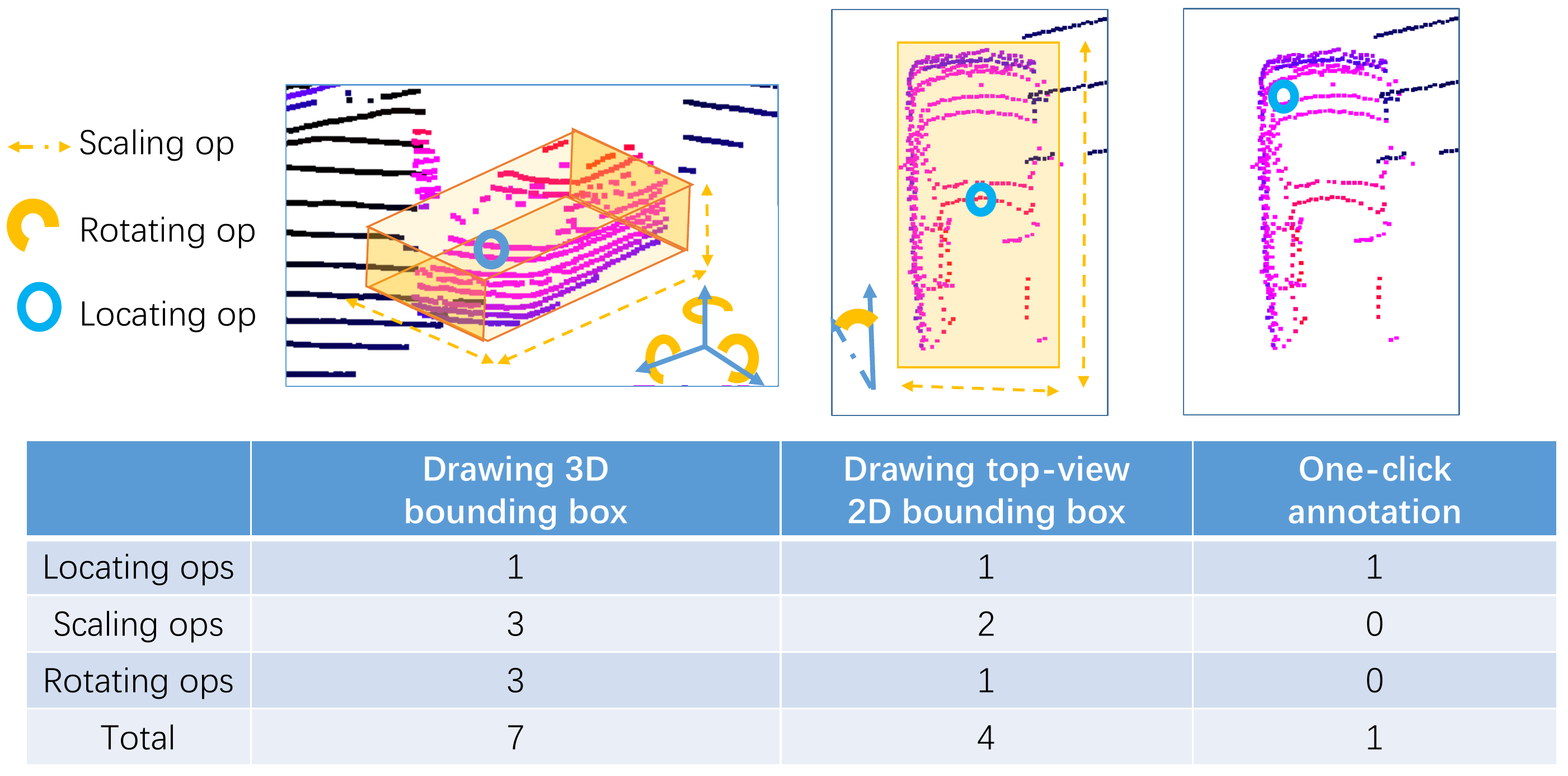}
    \caption{A comparison of drawing a 3D bounding box, a top-view 2D bounding box, and one-click annotation.}
    \label{fig:ops-comparison}
\end{figure}

In this section, we discuss how we simplify the annotation operation from drawing point-wise labels to drawing 3D bounding box, then to top-view 2D bounding boxes, and eventually to simply one-click annotation. A comparison of 3D bounding box, top-view 2D bounding box, and one-click annotation is illustrated in Fig.\ref{fig:ops-comparison}.

LiDAR-based perception can be formulated as a bounding box detection problem or a semantic segmentation problem. The former requires annotating 3D bounding boxes as shown in Fig.\ref{fig:challenges}(b), while the latter requires annotating point-wise labels. Naively annotating each point to obtain point-wise categorical labels is not feasible. Fortunately for 3D point cloud, point-wise labels can be obtained from 3D bounding boxes, as explained in \cite{wu2018squeezeseg}. For most of the road-objects we care about, their bounding boxes do not overlap in 3D space. As a result, point-wise labels can be converted from 3D bounding boxes, simply by treating each point inside a bounding box as part of the target object and therefore with the same the same categorical label. 

However, drawing 3D bounding boxes is still operationally complex. As illustrated in Fig. \ref{fig:ops-comparison}, ideally, drawing a 3D bounding box requires 1 operation to locate the object, 3 operations to scale the sides of the bounding box, and 3 rotations to adjust the orientation. For autonomous driving applications, what is more important is to locate the object from the top-view. Therefore, we can simplify a 3D bounding box to a top-view 2D bounding box, which can be determined by its 2D center position, 2D sizes of width and length, and its yaw angle. The operations needed to draw such a bounding box include 1 locating operation, 2 scaling operations, and 1 rotation, as illustrated in Fig. \ref{fig:ops-comparison}. 

\begin{figure}[h]
    \centering
    \includegraphics[width=\linewidth]{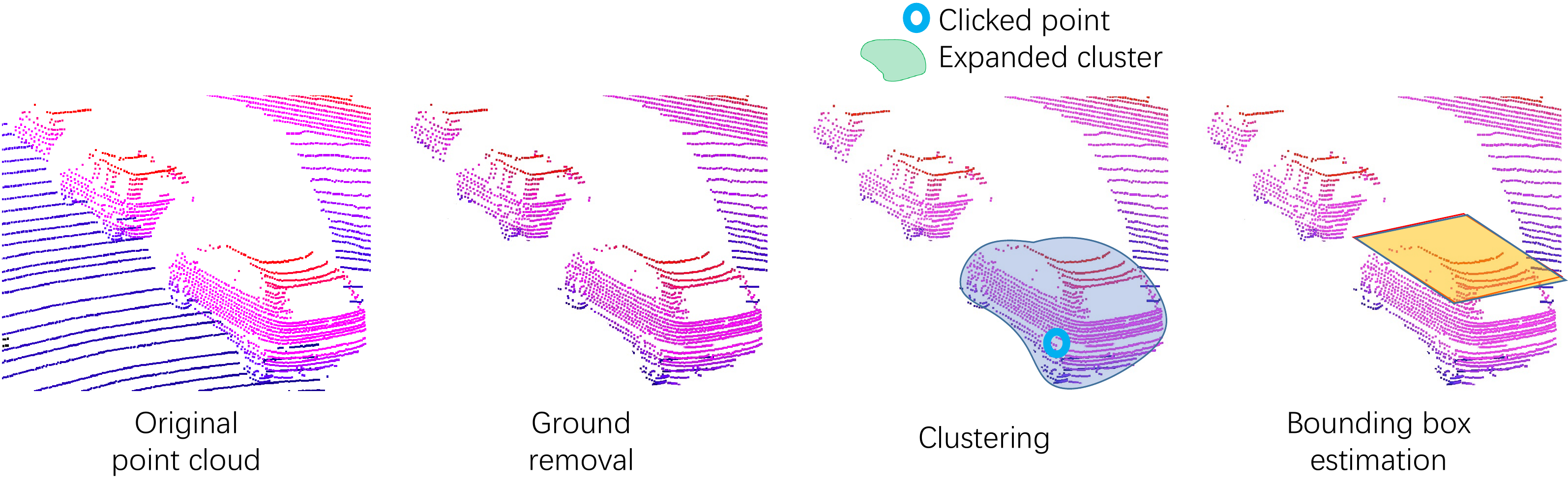}
    \caption{The one-click annotation pipeline of LATTE. For a given LiDAR point cloud, we first remove the ground. After an annotator clicks on one point on a target object, we use clustering algorithms to expand from the clicked point to the entire object. Finally, we estimate a top-view 2D bounding box for the object.}
    \label{fig:one-click}
\end{figure}

To further reduce the annotation complexity, we built one-click annotation -- human annotators only need to click on one point on the target object, as illustrated in Fig. \ref{fig:ops-comparison}. After the locating operations by human annotators, we automatically apply clustering algorithms to find all the points for the target object, from which we estimate a top-view 2D bounding box for the target object. Then, human annotators only need to adjust the automatically generated bounding box if it does not fit the object perfectly. Our one-click annotation is summarized in Fig. \ref{fig:one-click}. It mainly contains three steps: ground removal, clustering, and bounding box estimation.

\textbf{Ground removal:} We model the ground as a segment of planes where each plane is characterized by a linear model: 
\[
 \mathbf{n}^{T}\mathbf{p} = d,
\]
where $\mathbf{n} = [a, b, c]^T$ is the normal vector,  $\mathbf{p} = [x, y, z]^T$ is a point at the plane, and $d$ is the distance to the ground. To determine the ground, we need to estimate the normal vector $n$ from the noisy LiDAR data. We initially estimate the normal vector by sampling a set of lowest points in the vertical ditection (z-direction). We denote the set as $G_0$ and compute the covariance matrix $C_0 \in \mathcal{R}^{3 \times 3}$:
\begin{equation*}
\begin{gathered}
   \bar{\mathbf{p}} = \frac{1}{|G_0|}\sum_{i=1}^{|G_0|}{\mathbf{p}_i}, \\
    C_0 = \sum_{i=1}^{|G_0|}{(\mathbf{p}_i - \bar{\mathbf{p}})(\mathbf{p}_i - \bar{\mathbf{p}})^T}.
\end{gathered}
\end{equation*}

where $\bar{\mathbf{p}}$ is the mean of the points in $G_0$ and the covariance matrix $C_0$ represents how spread out the points in $G_0$ are. We analyze the direction of the dispersion by computing its singular value decomposition (SVD). The first two singular vectors corresponding to the two largest singular values represent the span of the plane. The singular vector corresponding to the smallest singular value is a good approximation for the normal because the variance in the direction of the normal is smallest among all directions. 

After computing the normal vector, we now have an updated estimate of the ground plane. With our estimated plane we resample the ground points by their distance and iteratively update the normal vector:
\[
G_{k} = \{\mathbf{p} : |\mathbf{n}_{k-1}^T \cdot \mathbf{p}| < thresh\}
\]
where $|\mathbf{n}_{k-1}^T \cdot \mathbf{p}|$ is the distance between point $p$ and the plane whose normal vector at iteration $k-1$ is $\mathbf{n}_{k-1}$. 
The normal vector approximation and ground sampling are repeated until the segmentation converges or for a fixed number of iterations. 

\textbf{Clustering:}
After removing the ground we find the nearest cluster to the point on which the human annotator clicks. The clustering algorithm is based on density-based spatial clustering of applications with noise (DBSCAN) \cite{Ester:1996:DAD:3001460.3001507} and is described in Algorithm \ref{alg:cluster}, where FindNeighbor(p, X, $\epsilon$) finds the  neigbors in X that are $\epsilon$-close to p.

\begin{algorithm}
\SetAlgoLined
\KwIn{seed $s \in \mathcal{R}^{3}$, point cloud $P \in \mathcal{R}^{n \times 3}$, distance threshold $\epsilon$}
\KwOut{cluster $C \in \mathcal{R}^{m \times 3}$}
 seen = $\emptyset$, initialize Q\;
 Q.push(seed)\;
 \While{Q not empty}{
  neighbors = FindNeighbors(Q.pop(), P, $\epsilon$)\; 
  \For{neighbor in neighbors}{
   \If{neighbor not in seen}{
    seen.add(neighbor)\;
    Q.push(neighbor)\;
   }
  }
  }
  return seen\;
 \caption{Clustering Algorithm}
 \label{alg:cluster}
\end{algorithm}

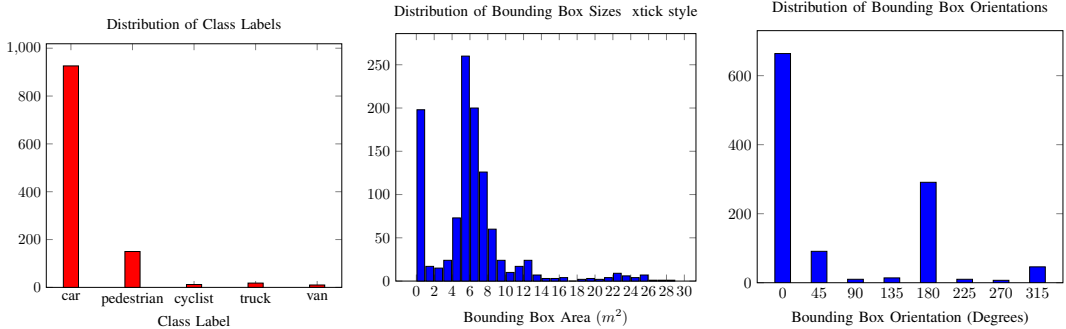
\begin{figure*}
\centering
\subfigure{\resizebox{0.26\linewidth}{!}{
\centering
\begin{tikzpicture}
\begin{axis}[
    title={Distribution of Class Labels},
    symbolic x coords={car, pedestrian, cyclist, truck, van},
    ymin=0,
    xlabel=Class Label]
]
\addplot[ybar, fill=red]
	coordinates {(car, 926) (pedestrian, 150) (cyclist, 12) (truck, 18) (van, 10)};
\end{axis}
\end{tikzpicture}
}}
\subfigure{\resizebox{0.26\linewidth}{!}{
\centering
\begin{tikzpicture}
\begin{axis}[
    title={Distribution of Bounding Box Sizes }
    xtick style={draw=none},
    xtick={0,2,...,30},
    ymin=0,
    bar width=5pt,
    xlabel=Bounding Box Area $(m^2)$]
]
\addplot[ybar, fill=blue]
	coordinates {(.5, 198) (1.5, 17) (2.5, 15) (3.5, 24) (4.5, 73) (5.5, 260) (6.5, 200) (7.5, 126) (8.5, 60) (9.5, 24) (10.5, 10) (11.5, 17) (12.5, 24) (13.5, 7) (14.5, 3) (15.5, 3) (16.5, 4) (17.5, 0) (18.5, 2) (19.5, 3) (20.5, 2) (21.5, 4) (22.5, 9) (23.5, 6) (24.5, 4) (25.5, 7) (26.5, 1) (27.5, 1) (28.5, 1)};
\end{axis}
\end{tikzpicture}
}}
\subfigure{\resizebox{0.26\linewidth}{!}{
\centering
\begin{tikzpicture}
\begin{axis}[
    title={Distribution of Bounding Box Orientations},
    xtick style={draw=none},
    xtick={0,45,...,360},
    ymin=0,
    xlabel=Bounding Box Orientation (Degrees)]
]
\addplot[ybar, fill=blue]
	coordinates {(0, 664) (45, 91) (90, 10) (135, 14) (180, 291) (225, 10) (270, 7) (315, 46)};
\end{axis}
\end{tikzpicture}
}}
\caption{Distribution of bounding boxes by class, box area, and orientation in our test benchmark.}
\label{fig:stats}
\end{figure*}

Since LiDAR point clouds can contain a large number of points (approximately 100,000 points per frame for Velodyne LiDAR), we perform pruning and downsampling in order to make one-click annotation efficient. Fig. \ref{fig:stats} shows that the distribution of bounding box sizes is concentrated around 6$m^2$ which is the size of a typical car. Therefore we can assume an upper bound on the dimensions of an object and appropriately prune the point cloud.  

\textbf{Bounding box estimation:} After we find the cluster, we use a search-based rectangle fitting \cite{Zhang-2017-26536} to estimate bounding boxes. Other methods, such as PCA based ones, can also be plugged into LATTE. To have the optimal rectangle fitting for a cluster, we need to know the appropriate heading of the rectangle. Ideally, the rectangle can be found by solving the following optimization problem:
\begin{equation*}
 \begin{aligned}
  & \argmin_{\theta, U, V, c_1, c_2 } & &\sum_{i \in |U|}{(x_i \cos\theta + y_i \sin\theta - c_1)^2} +  \\
  & & &\sum_{j \in |V|}{(-x_j \sin\theta + y_j \cos\theta - c_2)^2}, \\ 
  & \text{subject to } & & U \cup V = G, U \cap V = \emptyset, \\ 
 \end{aligned}
\end{equation*}
which aims to partition observed points in $G$ into two mutually exclusive groups $U$ and $V$ depending on which edges they are closer to. Points in $U$ are closer to the edge $x\cos\theta + y\sin\theta - c_1 $, and points in $V$ are closer to $-x\sin\theta + y\cos\theta - c_2$. 

Due to the combinatorial nature of the problem, it is infeasible to solve it exactly. To solve this problem approximately and efficiently, we use search-based rectangle fitting algorithm \cite{Zhang-2017-26536} that searches headings and projects the points in the cluster to two perpendicular edges. It searches the optimal heading to minimize a loss function as:
\begin{equation*}
    \theta^{*} = \argmin_{\theta \in [0, \pi]} L(G\mathbf{e}_{\theta,1}, G \mathbf{e}_{\theta,2}),
\end{equation*}
where $G \in \mathcal{R}^{n\times 2}$ denotes a matrix where each row contains the $(x, y)$-coordinate of a point. $\mathbf{e}_{\theta,1} = [\cos\theta, \sin\theta]^T, \mathbf{e}_{\theta,2} = [-\sin\theta, \cos\theta]^T$ are orthogonal unit vectors representing the directions of two perpendicular edges. The loss function $L(\cdot, \cdot)$ is defined as the following. We denote $\mathbf{c}_1 = G\mathbf{e}_{\theta,1}, \mathbf{c}_2 = G\mathbf{e}_{\theta,2}$, which represent projection of points to $\mathbf{e}_{\theta,1}, \mathbf{e}_{\theta,2}$. Then the distances from points in $G$ to the closer edge is computed as 
\begin{equation*}
    \begin{gathered}
        \mathbf{d}_1 = \argmin_{\mathbf{v} \in \{\mathbf{c}_1 - \min \{\mathbf{c}_1\}, \mathbf{c}_1 - \max \{\mathbf{c}_1\}\}} \|\mathbf{v}\|_2, \\
        \mathbf{d}_2 = \argmin_{\mathbf{v} \in \{\mathbf{c}_2 - \min \{\mathbf{c}_2\}, \mathbf{c}_2 - \max \{\mathbf{c}_2\}\}} \|\mathbf{v}\|_2.
    \end{gathered}
\end{equation*}
We can then divide all the points to two groups according to above distances and compute the loss function as
\begin{equation*}
 \begin{aligned}
  L(G\mathbf{e}_{\theta, 1}, G\mathbf{e}_{\theta, 2}) = & - \mathrm{Var}(\{\mathbf{d}_{1,i} : \mathbf{d}_{1,i} < \mathbf{d}_{2,i}\}) \\ & - \mathrm{Var}(\{\mathbf{d}_{2,i} : \mathbf{d}_{2,i} < \mathbf{d}_{1,i}\})
\end{aligned}
\end{equation*}
where $\mathbf{d}_{1, i}$ is the $i$-th element of $\mathbf{d}_{1}$, and similar for $\mathbf{d}_{2, i}$. $\mathrm{Var}(\cdot)$ computes the variance of a set of values. After solving for the optimal heading $\theta^*$, the following equations are used to compute the four edges of the rectangle: 
\begin{equation*}
 \begin{aligned}
\text{Edge 1:} & x\cos\theta^* + y\sin\theta^* - \min\{\mathbf{c}_{1}^*\} = 0 \\
\text{Edge 2:} & x\cos\theta^* + y\sin\theta^* - \max\{\mathbf{c}_{1}^*\} = 0 \\
\text{Edge 3:} & -x\sin\theta^* + y\cos\theta^* - \min\{\mathbf{c}_{2}^*\} = 0 \\
\text{Edge 4:} & -x\sin\theta^* + y\cos\theta^* - \max\{\mathbf{c}_{2}^*\} = 0
\end{aligned}
\end{equation*}
where  $\mathbf{c}_1^* = G\mathbf{e}_{\theta^{*},1},
\mathbf{c}_2^* = G\mathbf{e}_{\theta^{*},2}$.

\subsection{Tracking}
\begin{figure}[h]
    \centering
    \includegraphics[width=0.75\linewidth]{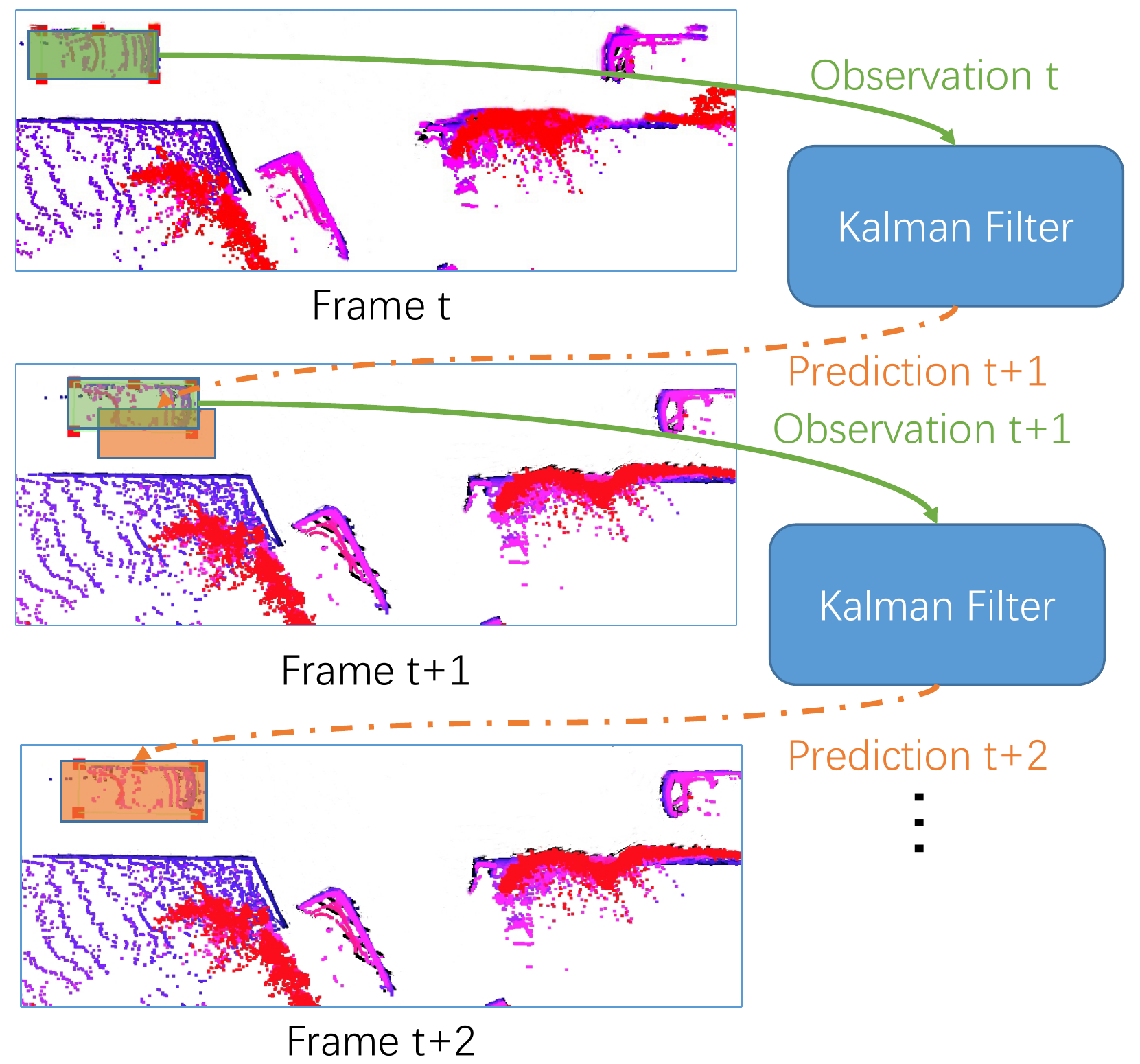}
    \caption{Tracking pipeline of LATTE. Annotators label a bounding box in the initial frame. Next, we use Kalman filtering to predict the center position of the bounding box at the next frame. Human annotators then adjust the bounding box, and we use the new center position as a new observation to update the Kalman filter.}
    \label{fig:tracking}
\end{figure}

To accelerate annotation on sequences, we integrate tracking to LATTE such that annotations from one frame can be transferred to subsequent ones, as illustrated in Fig. \ref{fig:tracking}.

LATTE is constructed in a modular way such that it can support different tracking algorithms, but we adopt Kalman filtering \cite{Welch:1995:IKF:897831} in our implementation. We use Kalman filtering to track the bounding box center of a target object. Human annotators need to label the first frame of a sequence. Next, our algorithm predicts the centers of bounding boxes for the next frame. For non-rigid objects (such as pedestrians), their bounding boxes do not have fixed shapes. Therefore, we re-estimate the bounding boxes using a similar algorithm as the one-click annotation, as described in section \ref{sec:one-click}. For rigid objects such as cars, their bounding boxes should not change over time, so we only estimate the yaw angle. The predicted bounding box is displayed at the next frame, and the human annotator can simply make adjustments to the bounding boxes. The adjusted bounding boxes then serve as observations in the Kalman update step. 

Formally, we define the state vector of a bounding box at frame $k$ as $x_k = [p_x, p_y, v_x, v_y, a_x, a_y]^T$, where $p_x$ and $p_y$ are the coordinates for the center of the bounding box at frame $k$. $v_x$, $v_y$ and $a_x$, $a_y$ represent the velocity and acceleration of the center, respectively. A human annotates the first frame in the sequence, and we can obtain the initial values for the center position. The initial velocity and acceleration values are left to be zero. Next, we predict the center coordinates at the next frame as 
\begin{equation*}
\begin{gathered}
    \mathbf{\hat{x}}_{k|k-1} = F\mathbf{\hat{x}}_{k-1|k-1}, \\
    P_{k|k-1}=F P_{k-1|k-1} F^{T}+Q,
\end{gathered}
\end{equation*}
where $F \in \mathcal{R}^{6 \times 6}$ represents the state transition model. We assume a constant acceleration model and define $F$ as 
\[
\mathbf{F} = \begin{bmatrix} 
1 & 0 & \Delta t & 0 & \frac{1}{2} \Delta t^2 & 0\\
0 & 1 & 0 & \Delta t & 0 & \frac{1}{2} \Delta t^2\\
0 & 0 & 1 & 0 & \Delta t & 0\\ 
0 & 0 & 0 & 1 & 0 & \Delta t\\ 
0 & 0 & 0 & 0 & 1 & 0\\
0 & 0 & 0 & 0 & 0 & 1\\
\end{bmatrix},
\]
where $\Delta t$ is the sampling interval of the sensor. $Q \in \mathcal{R}^{6\times 6}$ represents the process noise covariance matrix and is modeled as $Q = \text{diag}(n_x, n_y, n_{v_x}, n_{v_y}, n_{a_x}, n_{a_y})$, where the coefficients are tuned by trial-and-error. The values are based their on the units of the state vector ($m$, $m/s$, $m/s^2$) and their uncertainty level. Since we can directly observe center coordinates, we have higher certainty for $n_x$ and $n_y$ than others. $\hat{x}_{k-1|k-1}$ represents the a posteriori state estimate at time $k-1$ given observations up to and including at time k. $P_{k-1|k-1} \in \mathcal{R}^{6\times 6}$ represents the a posteriori error covariance matrix, and the initial value $\mathbf{P}_{0|0}$ is estimated empirically by computing the error covariance matrix on a sample of 100 tracking objects. 

Based on the center position prediction, we then estimate the bounding box at frame $k$ and ask the human annotator to adjust it. The adjusted bounding box provides us an observation of the new center coordinates $\mathbf{z}_{k} = [p_{x,k}, p_{y,k}]^T$, which we use to update the Kalman filter as 
\begin{equation*}
    \begin{gathered}
    \mathbf{\tilde{y}}_{k} =  \mathbf{z}_{k} - H\mathbf{\hat{x}}_{k|k-1}, \\
    K_k = P_{k|k-1}H^{T} (R + H P_{k|k-1}H^T)^{-1}, \\
    \mathbf{\hat{x}}_{k|k} = \mathbf{\hat{x}}_{k|k-1} + K_k \mathbf{\tilde{y}}_{k},  \\
    P_{k|k} = (I - K_k H)P_{k|k-1}(I - K_k H)^T + K_k R K_k^T,
    \end{gathered}
\end{equation*}
where $H \in \mathcal{R}^{2\times 6}$ is the observation model. Since we only observe the center position through, we define
\[
H = \begin{bmatrix} 
1 & 0 & 0 & 0 & 0 & 0\\
0 & 1 & 0 & 0 & 0 & 0\\
\end{bmatrix}.
\]
$R \in \mathcal{R}^{2\times 2}$ is the covariance of the observation noise and is defined as $R = \text{diag}(\Delta x, \Delta y)$ where $\Delta x$ and $\Delta y$ are the resolution of the LiDAR scanner in the x and y direction. We iteratively apply Kalman filtering from the first frame to the end, as shown in Fig. \ref{fig:tracking}.

\section{Experiments}
\label{sec:experiment}
\subsection{Experiment Setup}
Providing precise quantitative measurements of productivity improvements using human subjects is quite time consuming and complicated \cite{lazar2017research}. Nevertheless, while we feel that the productivity advantages of features such as "one-click annotation" are obvious, we wanted to provide some estimate of the productivity improvements that LATTE provided. So, to estimate productivity we simply measured the time and operation count used by volunteers using LATTE to annotate LiDAR point cloud data from the KITTI dataset \cite{KITTI}. The KITTI dataset involves eight object categories and includes 3D Velodyne point cloud data, accompanying color images, GPS/IMU data, 3D object tracklet labels, and camera-to-Velodyne calibration data. We randomly selected 30 sequences of LiDAR data, where each sequence contains five frames. This test benchmark contains a total amount of 1,116 instances. More detailed analysis of object statistics can be found in Fig. \ref{fig:stats}. 

We asked nine volunteers to annotate these frames using LATTE with all three features (sensor fusion, one-click annotation, and tracking). We divide the data and volunteers such that each feature is evaluated on the entire dataset. In other words each frame is annotated using each feature so that every feature is evaluated on the same frames. This is to ensure that we are testing each feature on the same data. 

The annotators were asked to draw bounding boxes for instances for which they feel confident, for example instances of vehicles where at least two complete edges are visible. Instances that are far away tend to be sparse or occluded and are therefore not annotated. We only evaluate objects whose bounding box intersects with a ground truth bounding box. To form a baseline for comparison, we asked volunteers to draw top-view 2D bounding boxes on the test LiDAR point cloud without using LATTE's advanced features. To further evaluate the efficacy of each component, we also asked volunteers to use only one of the three features for annotation. Each volunteer annotates a sequence of 5 frames with one feature a time. We also asked the volunteers to annotate with a fully-featured version of the tool. In order to eliminate the case where annotation efficiency is improved solely due to the fact that the annotator has seen the frame before, each frame is seen only once by each volunteer. Therefore we do not falsely attribute an improvement in efficiency to the feature we are testing. 

\subsection{Metrics}
To evaluate the accuracy of annotations, we used our own ground truth instead of using the bounding boxes from KITTI dataset. This is because bounding boxes provided by KITTI do not include all instances in a scene, particularly the ones that are behind the drive. Therefore, we asked an expert human annotator to provide high-quality annotations as ground truth. We measure the \textbf{intersection-over-union (IoU)} between an annotated bounding box and the ground truth as the accuracy metric per instance, and we report the IoU averaged over all the instances annotated by all of the volunteers. Note that comparing our ground truth with KITTI's bounding boxes, we see 86.0\% average IoU. 

\begin{figure}[h]
    \centering
    \includegraphics[width=\linewidth]{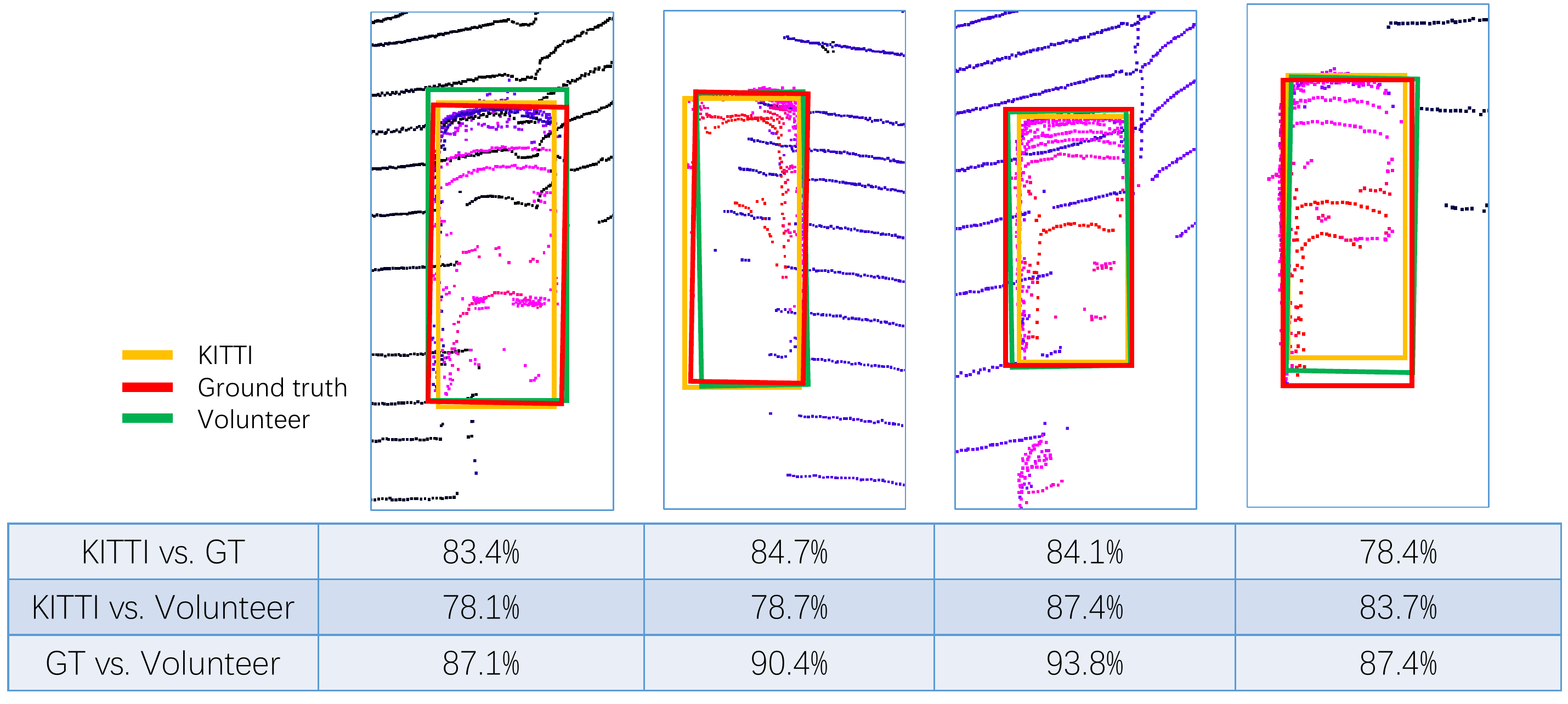}
    \caption{Visualization of bounding box annotations from our volunteers, our ground truth, and KITTI. The IoU between each pair listed. }
    \label{fig:iou-viz}
\end{figure}

To better understand the typical agreement between two human annotators, we ask different volunteers to use LATTE to label the same frames and instances, and compute the pair-wise IoU agreement per instance. Among 452 pairs of annotations on 132 instances, the average IoU is 84.5\% with a standard deviation of 8.74\%. This serves as a reference for considering other IoU results. In addition, we select a few samples of bounding box annotations and compare them with our ground truth and KITTI's annotation in Fig. \ref{fig:iou-viz}. As we can see, the IoU between each pairs ranges from 78.1\% to 93.8\%, but the bounding boxes are very similar to each other. 

Another factor of accuracy is the correct identification of objects of interest. To evaluate this, we measure the \textbf{instance-level precision and recall}. An annotation has to have more than 50\% of the IoU with a ground truth box in order to be considered a true positive. This is different from the IoU above since IoU is counted on objects that are both annotated by a volunteer and the ground truth. We report the IoU result in Table \ref{table:efficiency_results} and the precision and recall in Table \ref{tab:prec-recall}. 

To evaluate efficiency, we record the \textbf{time} spent on annotation per instance. In addition, we also measure the \textbf{number of annotation operations}, which is defined as bounding box adjustments (resizing, rotation, translation) and assigning classes. The efficiency result is reported in Table \ref{table:efficiency_results}. 

\subsection{Results}
\begin{table}[h]
\begin{threeparttable}
\caption{Accuracy and efficiency of LATTE.}
\label{table:efficiency_results}
\centering
\begin{tabular}{c|c|c|c|c}
\hline
Method               & IoU (\%) & \begin{tabular}[c]{@{}c@{}}IoU(\%)\\ w/ KITTI\end{tabular} & Time (s) & \#ops \\ \hline
Ground Truth         & 100.0    & 86.0                                                       & -        & -     \\ \hline
Baseline             & 85.5     & 82.3                                                       & 9.51      & 3.76   \\
Sensor fusion        & 86.3     & 82.5                                                       & 3.88      & 2.88   \\
One-click annotation & 86.2     & 82.9                                                       & 2.55      & 1.29   \\
Tracking             & 86.4     & 83.5                                                       & 2.41      & 1.53   \\
Full features        & 87.5     & 84.7                                                       & 1.53      & 1.02   \\ \hline
\end{tabular}
\begin{tablenotes}
\item The accuracy and efficiency of LATTE is listed. The  ``IoU'' column shows the average IoU of annotation vs. our ground truth. The ``IoU w/ KITTI'' column shows the average IoU of annotations vs. the KITTI ground truth. The ``Time'' column shows the average time spent on annotating one instance. The ``\#ops'' column shows the average number of operations spent on annotating one instance. As a reference for the IoU result, the average pair-wise IoU from different annotators is 84.5\% with a standard deviation of 8.74\%.  
\end{tablenotes}
\end{threeparttable}
\end{table}

\begin{table}[h]
\centering
\caption{Instance-level Precision \& Recall of LATTE}
\label{tab:prec-recall}
\begin{tabular}{c|c|c}
\hline
Method & Precision (\%) & Recall (\%) \\ \hline
Baseline & 69.9 & 82.9 \\
Sensor fusion & 83.9 & 85.0 \\
One-click annotation & 78.8 & 84.8 \\
Tracking & 91.8 & 85.2 \\
Full features & 93.5 & 85.1 \\ 
\hline
\end{tabular}
\end{table}

\textbf{Baseline}: Since there are no other open-source tools with similar functionality, we compare LATTE to a stripped down version with all of the new features removed. The volunteers are asked to manually annotate all instances by drawing top-view bounding boxes, as illustrated in Fig. \ref{fig:ops-comparison}. From Table \ref{table:efficiency_results}, we can see that it takes an average of 9.51s, and 3.76 operations to annotate one instance. Despite the longer time, the label quality is the poorest among all variations. The IoU with the ground truth is just 85.5\%, the precision is 69.9\%, and the recall is 82.9\%.

\textbf{Full features:} To test LATTE with all of the features, we ask the volunteers to use the following workflow. For the first frame of a sequence, we use sensor fusion to highlight objects of interest and ask human annotators to use one-click annotation to draw annotations and make necessary adjustments. After volunteers are confident with first frame, they can move on to the next frame, where tracking will generate bounding box proposals for volunteers to verify and adjust. The full features show a 6.2 times speed-up in annotation time and 3.7 times reduction in number of operations while achieving higher IoU at 87.5\% over the baseline's 85.5\%, higher recall (85.1\% vs. 82.9\%) and significantly higher precision (93.5\% vs. 69.9\%). We visualize some samples of annotations and compare them with the ground truth and KITTI dataset's annotations in Fig. \ref{fig:iou-viz}. 

\textbf{Ablation study}: To study the effectiveness of each of the proposed features, we ask volunteers to use a variation of LATTE with only one feature enabled. When volunteers are only allowed to use the \textbf{sensor fusion} feature, we are able to achieve a 2.4 times  speed-up in annotation time compared to the baseline while achieving higher IoU (+0.8\%), recall (+2.1\%), and significantly higher precision (+14.0\%). In addition, the number of operations per instance on average reduces by 0.9, nearly an entire operation. This supports our claim that sensor fusion helps human annotators to better recognize objects from point clouds. With only \textbf{one-click annotation}, our method shows a 3.8 times speed-up in annotation time and 2.9 reduction in the number of annotation operations while achieving higher IoU, precision and recall. With only \textbf{tracking},  we show a speed-up of 4.74x while achieving better annotation agreement than the baseline. This is largely due to the fact that tracking saved redundant annotations on sequential frames.

\section{CONCLUSIONS}
Efficiently annotating LiDAR point clouds at scale is crucial for the development of LiDAR-based detection and autonomous driving. However, annotating LiDAR point clouds is difficult due to the challenges of low resolution, complex annotating operations, and sequential correlation. To solve these problems, we propose LATTE, an open-sourced LiDAR annotation tool that features sensor-fusion, one-click annotation, and tracking. Based on our own experiments we estimate that LATTE achieves a 6.2x speedup compared with baseline annotation tools and delivers better label quality with 23.6\% and 2.2\% higher instance-level precision and recall, and 2.0\% higher bounding box IoU.  

\addtolength{\textheight}{-12cm}   




\bibliography{bibliography.bib}

\begin{thebibliography}{10}
\providecommand{\url}[1]{#1}
\csname url@rmstyle\endcsname
\providecommand{\newblock}{\relax}
\providecommand{\bibinfo}[2]{#2}
\providecommand\BIBentrySTDinterwordspacing{\spaceskip=0pt\relax}
\providecommand\BIBentryALTinterwordstretchfactor{4}
\providecommand\BIBentryALTinterwordspacing{\spaceskip=\fontdimen2\font plus
\BIBentryALTinterwordstretchfactor\fontdimen3\font minus
  \fontdimen4\font\relax}
\providecommand\BIBforeignlanguage[2]{{%
\expandafter\ifx\csname l@#1\endcsname\relax
\typeout{** WARNING: IEEEtran.bst: No hyphenation pattern has been}%
\typeout{** loaded for the language `#1'. Using the pattern for}%
\typeout{** the default language instead.}%
\else
\language=\csname l@#1\endcsname
\fi
#2}}

\bibitem{wu2018squeezeseg}
B.~Wu, A.~Wan, X.~Yue, and K.~Keutzer, ``Squeezeseg: Convolutional neural nets
  with recurrent crf for real-time road-object segmentation from 3d lidar point
  cloud,'' in \emph{2018 IEEE International Conference on Robotics and
  Automation (ICRA)}.\hskip 1em plus 0.5em minus 0.4em\relax IEEE, 2018, pp.
  1887--1893.

\bibitem{wu2018squeezesegv2}
B.~Wu, X.~Zhou, S.~Zhao, X.~Yue, and K.~Keutzer, ``Squeezesegv2: Improved model
  structure and unsupervised domain adaptation for road-object segmentation
  from a lidar point cloud,'' \emph{arXiv preprint arXiv:1809.08495}, 2018.

\bibitem{qi2017frustum}
C.~R. Qi, W.~Liu, C.~Wu, H.~Su, and L.~J. Guibas, ``Frustum pointnets for 3d
  object detection from rgb-d data,'' \emph{arXiv preprint arXiv:1711.08488},
  2017.

\bibitem{LiDARDet}
B.~Li, T.~Zhang, and T.~Xia, ``Vehicle detection from 3d lidar using fully
  convolutional network,'' \emph{arXiv preprint arXiv:1608.07916}, 2016.

\bibitem{KITTI}
A.~Geiger, P.~Lenz, and R.~Urtasun, ``Are we ready for autonomous driving? the
  kitti vision benchmark suite,'' in \emph{Computer Vision and Pattern
  Recognition (CVPR), 2012 IEEE Conference on}.\hskip 1em plus 0.5em minus
  0.4em\relax IEEE, 2012, pp. 3354--3361.

\bibitem{huang2018apolloscape}
X.~Huang, X.~Cheng, Q.~Geng, B.~Cao, D.~Zhou, P.~Wang, Y.~Lin, and R.~Yang,
  ``The apolloscape dataset for autonomous driving,'' in \emph{Proceedings of
  the IEEE Conference on Computer Vision and Pattern Recognition Workshops},
  2018, pp. 954--960.

\bibitem{LiDARSegICRA2012}
B.~Douillard, J.~Underwood, N.~Kuntz, V.~Vlaskine, A.~Quadros, P.~Morton, and
  A.~Frenkel, ``On the segmentation of 3d lidar point clouds,'' in
  \emph{Robotics and Automation (ICRA), 2011 IEEE International Conference
  on}.\hskip 1em plus 0.5em minus 0.4em\relax IEEE, 2011, pp. 2798--2805.

\bibitem{himmelsbach2008lidar}
M.~Himmelsbach, A.~Mueller, T.~L{\"u}ttel, and H.-J. W{\"u}nsche, ``Lidar-based
  3d object perception,'' in \emph{Proceedings of 1st international workshop on
  cognition for technical systems}, vol.~1, 2008.

\bibitem{wang2012could}
D.~Z. Wang, I.~Posner, and P.~Newman, ``What could move? finding cars,
  pedestrians and bicyclists in 3d laser data,'' in \emph{Robotics and
  Automation (ICRA), 2012 IEEE International Conference on}.\hskip 1em plus
  0.5em minus 0.4em\relax IEEE, 2012, pp. 4038--4044.

\bibitem{qi2017pointnet}
C.~R. Qi, H.~Su, K.~Mo, and L.~J. Guibas, ``Pointnet: Deep learning on point
  sets for 3d classification and segmentation,'' in \emph{CVPR}, 2017, pp.
  77--85.

\bibitem{qi2017pointnet++}
C.~R. Qi, L.~Yi, H.~Su, and L.~J. Guibas, ``Pointnet++: Deep hierarchical
  feature learning on point sets in a metric space,'' in \emph{NIPS}, 2017, pp.
  5099--5108.

\bibitem{dutta2016via}
A.~Dutta, A.~Gupta, and A.~Zissermann, ``{VGG} image annotator ({VIA}),''
  \url{http://www.robots.ox.ac.uk/~vgg/software/via/}, 2016.

\bibitem{castrejon2017annotating}
L.~Castrejon, K.~Kundu, R.~Urtasun, and S.~Fidler, ``Annotating object
  instances with a polygon-rnn,'' in \emph{Proceedings of the IEEE Conference
  on Computer Vision and Pattern Recognition}, 2017, pp. 5230--5238.

\bibitem{acuna2018efficient}
D.~Acuna, H.~Ling, A.~Kar, and S.~Fidler, ``Efficient interactive annotation of
  segmentation datasets with polygon-rnn++,'' 2018.

\bibitem{vatic}
\BIBentryALTinterwordspacing
C.~Vondrick, D.~Patterson, and D.~Ramanan, ``Efficiently scaling up
  crowdsourced video annotation,'' \emph{International Journal of Computer
  Vision}, pp. 1--21, 10.1007/s11263-012-0564-1. [Online]. Available:
  \url{http://dx.doi.org/10.1007/s11263-012-0564-1}
\BIBentrySTDinterwordspacing

\bibitem{yu2018bdd100k}
F.~Yu, W.~Xian, Y.~Chen, F.~Liu, M.~Liao, V.~Madhavan, and T.~Darrell,
  ``Bdd100k: A diverse driving video database with scalable annotation
  tooling,'' \emph{arXiv preprint arXiv:1805.04687}, 2018.

\bibitem{piewak2018boosting}
F.~Piewak, P.~Pinggera, M.~Schafer, D.~Peter, B.~Schwarz, N.~Schneider,
  M.~Enzweiler, D.~Pfeiffer, and M.~Zollner, ``Boosting lidar-based semantic
  labeling by cross-modal training data generation,'' in \emph{Proceedings of
  the European Conference on Computer Vision (ECCV)}, 2018, pp. 0--0.

\bibitem{yue2018lidar}
X.~Yue, B.~Wu, S.~A. Seshia, K.~Keutzer, and A.~L. Sangiovanni-Vincentelli, ``A
  lidar point cloud generator: from a virtual world to autonomous driving,'' in
  \emph{Proceedings of the 2018 ACM on International Conference on Multimedia
  Retrieval}.\hskip 1em plus 0.5em minus 0.4em\relax ACM, 2018, pp. 458--464.

\bibitem{Dosovitskiy17}
A.~Dosovitskiy, G.~Ros, F.~Codevilla, A.~Lopez, and V.~Koltun, ``{CARLA}: {An}
  open urban driving simulator,'' in \emph{Proceedings of the 1st Annual
  Conference on Robot Learning}, 2017, pp. 1--16.

\bibitem{he2017mask}
K.~He, G.~Gkioxari, P.~Doll{\'a}r, and R.~Girshick, ``Mask r-cnn,'' in
  \emph{Proceedings of the IEEE international conference on computer vision},
  2017, pp. 2961--2969.

\bibitem{Ester:1996:DAD:3001460.3001507}
\BIBentryALTinterwordspacing
M.~Ester, H.-P. Kriegel, J.~Sander, and X.~Xu, ``A density-based algorithm for
  discovering clusters a density-based algorithm for discovering clusters in
  large spatial databases with noise,'' in \emph{Proceedings of the Second
  International Conference on Knowledge Discovery and Data Mining}, ser.
  KDD'96.\hskip 1em plus 0.5em minus 0.4em\relax AAAI Press, 1996, pp.
  226--231. [Online]. Available:
  \url{http://dl.acm.org/citation.cfm?id=3001460.3001507}
\BIBentrySTDinterwordspacing

\bibitem{Zhang-2017-26536}
X.~Zhang, W.~Xu, C.~Dong, and J.~M. Dolan, ``Efficient l-shape fitting for
  vehicle detection using laser scanners,'' in \emph{2017 IEEE Intelligent
  Vehicles Symposium}, June 2017.

\bibitem{Welch:1995:IKF:897831}
G.~Welch and G.~Bishop, ``An introduction to the kalman filter,'' Chapel Hill,
  NC, USA, Tech. Rep., 1995.

\bibitem{lazar2017research}
J.~Lazar, J.~H. Feng, and H.~Hochheiser, \emph{Research methods in
  human-computer interaction}.\hskip 1em plus 0.5em minus 0.4em\relax Morgan
  Kaufmann, 2017.

\end{thebibliography}
\bibliographystyle{IEEEtran}

\end{document}